\newcount\isdebug
\ifnum\isdebug=1

\else
\newcommand{\yourname}[1]{}
\fi

\newcommand{\del}[1]{#1}

\newcommand{\fmrs}{Foundation Model-enabled robotics }
\newcommand{\Fmrs}{Foundation Model-enabled robotics }
\newcommand{\FMRs}{Foundation Model-enabled Robotics }

\documentclass{article}
\usepackage{ijcai25}

\usepackage{times}
\usepackage{soul}
\usepackage{url}
\usepackage[hidelinks]{hyperref}
\usepackage[utf8]{inputenc}
\usepackage[small]{caption}
\usepackage{graphicx}
\usepackage{amsmath}
\usepackage{amsthm}
\usepackage{amsfonts} % Custom addition
\usepackage{booktabs}
\usepackage{algorithm}
\usepackage{algorithmic}
\usepackage[switch]{lineno}

\usepackage{setspace}
\usepackage{lipsum}
\let\OLDthebibliography\thebibliography
\renewcommand\thebibliography[1]{
  \OLDthebibliography{#1}
  \setlength{\parskip}{1pt}
  \setstretch{0.1}
}
\usepackage[edges]{forest}
\usepackage{graphicx}
\usepackage{tikz}
\usepackage{xcolor,colortbl}
\usepackage{multirow}
\usepackage{makecell}
\usepackage{makecell}
\usepackage{xspace,mfirstuc,tabulary}
\usepackage{booktabs}
\usepackage{graphicx}
\usepackage{bbding}
\usepackage{xcolor,colortbl}
\usepackage{multirow}
\usepackage{adjustbox}

\usepackage{pifont}
\newrobustcmd{\B}{\bfseries}
\newcommand*\colourcheck[1]{%
  \expandafter\newcommand\csname #1check\endcsname{\textcolor{#1}{\ding{52}}}%
}
\newcommand*\colourcross[1]{%
  \expandafter\newcommand\csname #1cross\endcsname{\textcolor{#1}{\ding{55}}}%
}
\colourcheck{green}
\colourcross{red}
\newcommand{\eg}{\emph{e.g.}}

\usepackage[most]{tcolorbox}
\newtcolorbox{highlighted}{
  colback=yellow!50!white, colframe=yellow!50!white, boxrule=0pt, sharp corners, left=0pt, right=0pt, top=0pt, bottom=0pt
}

\usepackage{color,soul}
\soulregister{\textcolor}{2}
\soulregister{\cite}7
\soulregister{\citep}7
\soulregister{\citet}7
\soulregister{\ref}7

\usepackage[switch]{lineno}
\sethlcolor{yellow}

\definecolor{lightcoral}{rgb}{0.94, 0.5, 0.5}
\definecolor{lightgreen}{rgb}{0.56, 0.93, 0.56}
\definecolor{harvestgold}{rgb}{0.98, 0.85, 0.40}
\definecolor{brightlavender}{rgb}{0.75, 0.58, 0.89}
\definecolor{capri}{rgb}{0.0, 0.75, 1.0}
\definecolor{carminepink}{rgb}{0.92, 0.3, 0.26}
\definecolor{celadon}{rgb}{0.67, 0.88, 0.69}
\definecolor{darkpastelgreen}{rgb}{0.01, 0.75, 0.24}

\definecolor{hidden-draw}{RGB}{205, 44, 36}
\definecolor{hidden-blue}{RGB}{194,232,247}
\definecolor{hidden-orange}{RGB}{243,202,120}
\definecolor{hidden-yellow}{RGB}{242,244,193}
\definecolor{tree-level-1}{RGB}{245,20,85}
\definecolor{tree-level-2}{RGB}{246,86,118}
\definecolor{tree-level-3}{RGB}{248,177,193}
\definecolor{tree-leaf}{RGB}{176,230,198}
\title{A Comprehensive Survey on Physical Risk Control \\ in the Era of \FMRs}

\newcommand\blfootnote[1]{%
  \begingroup
  \renewcommand\thefootnote{}\footnote{#1}%
  \addtocounter{footnote}{-1}%
  \endgroup
}

\author{
Takeshi Kojima$^{1\dagger}$\and
Yaonan Zhu$^{1\dagger}$\and
Yusuke Iwasawa$^{1\dagger}$\and
Toshinori Kitamura$^1$\and
\\
Gang Yan$^1$\and 
Shu Morikuni$^1$\and
Ryosuke Takanami$^1$\and
Alfredo Solano$^1$\and 
\\
Tatsuya Matsushima$^1$\and
Akiko Murakami$^2$\And
Yutaka Matsuo$^1$
\affiliations
$^1$The University of Tokyo\\
$^2$Japan AI Safety Institute\\
\emails
\{t.kojima, yaonan.zhu, iwasawa\}@weblab.t.u-tokyo.ac.jp
}

\begin{document}
\maketitle

%\blfootnote{}
\blfootnote{$^\dagger$ Corresponding Authors.}

% ==========================================
\begin{abstract}

% Recent robotic foundation models (RFMs) have achieved excellent improvements in general-purpose skills.
% While we could enjoy versatile benefits by replacing human labor with robots in the near future, we cannot completely avoid physical damage risks to surrounding humans or objects induced by RFMs.

Recent \fmrs (FMRs) display greatly improved general-purpose skills, enabling more adaptable automation than conventional robotics. Their ability to handle diverse tasks thus creates new opportunities to replace human labor. However, unlike general foundation models, FMRs interact with the physical world, where their actions directly affect the safety of humans and surrounding objects, requiring careful deployment and control.
Based on this proposition, our survey comprehensively summarizes robot control approaches to mitigate physical risks by covering all the lifespan of FMRs ranging from pre-deployment to post-accident stage.
Specifically, we broadly divide the timeline into the following three phases: (1) pre-deployment phase, (2) pre-incident phase, and (3) post-incident phase.
Throughout this survey, we find that there is much room to study (i) pre-incident risk mitigation strategies, (ii) research that assumes physical interaction with humans, and (iii) essential issues of foundation models themselves.
We hope that this survey will be a milestone in providing a high-resolution analysis of the physical risks of FMRs and their control, contributing to the realization of a good human-robot relationship.

%From this survey, we found that while there is an increasing attention to the safety control for FMRs, there is much room to accelerate research that assumes robot development in open world settings where humans and robots are frequently interacting with each other (e.g., red-teaming to promote safety, improving learning methods to strictly observe social rules, post-incident researches which aim to prevent more hazardous risks on-site immediately or for future improvement cycle).
%We conclude that while research on physical risk mitigation (1 and 2) is a widely recognized and evolving field, there is much room for technological improvement in the post-incident phase (3) for future work.
%We discuss that in addition to the technical aspects, social measures such as legislation or insurance schemes are also important to enhance aftercare of physical damage in practice.
%We hope that this study will be a milestone in providing a high resolution of the physical risks of FMRs and their control to realize a good human-robot relationship.

\end{abstract}
\section{Introduction}

\begin{figure}[t]
    \centering
    \includegraphics[width=1.0\columnwidth]{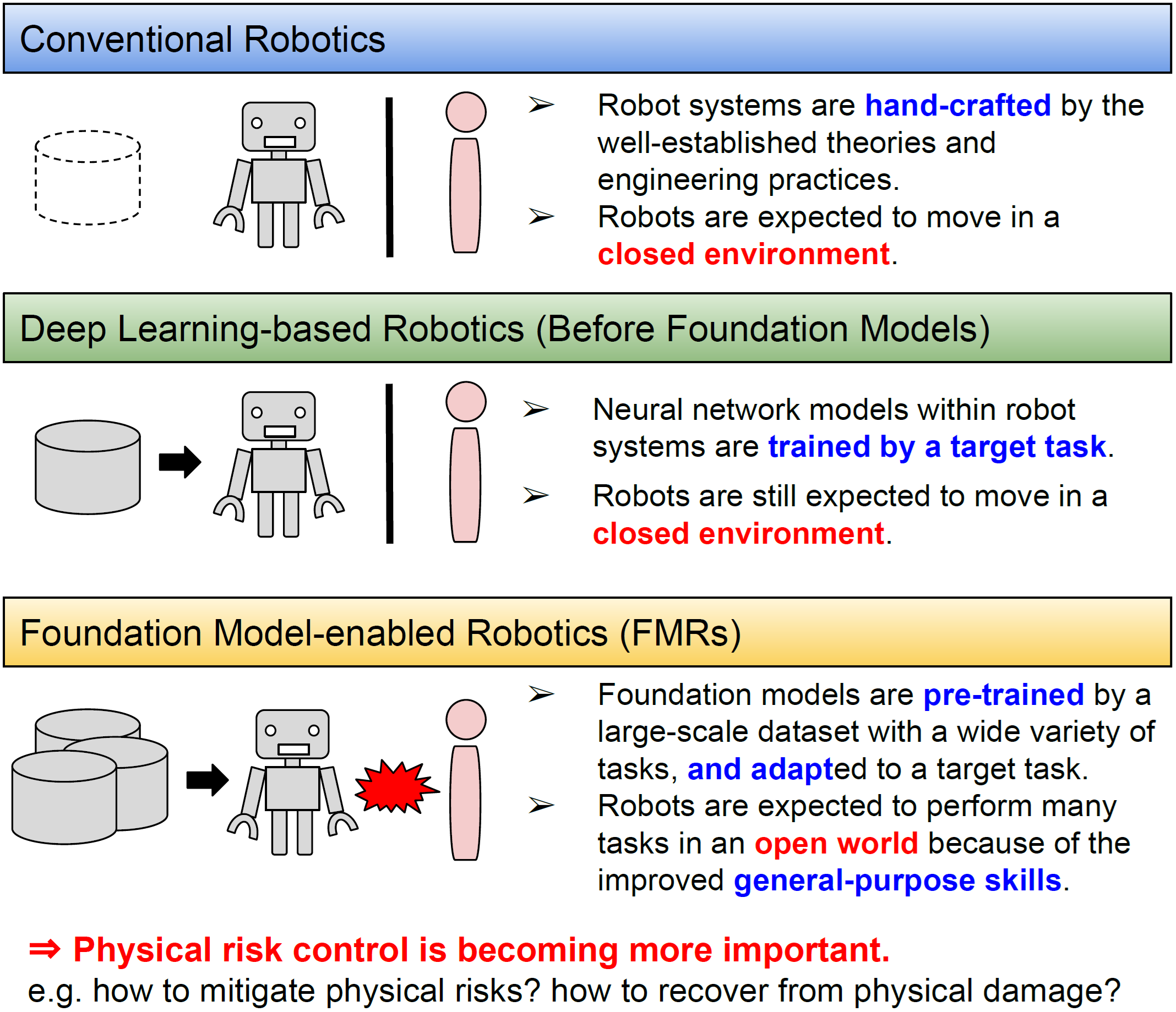}
    \caption{Motivation of our survey. Studies on conventional robotics and Deep Learning-based robotics (before foundation models) were mainly based on closed environments where physical risks are excluded or minimized by restricting the entry of humans and objects, e.g., inside a factory. \Fmrs (FMRs) are expected to be utilized in an open world where physical risks inevitably exist because humans and robots are always in close proximity and physically interacting, e.g. a restaurant. Hence physical risk control is becoming more important in the era of FMRs.
    %, e.g., how to mitigate physical risks? how to recover from physical damage?
    }
    \label{fig:motivation}
    %\vspace{-3mm}
\end{figure}

Since the emergence of foundation models, robotics has shown dramatic improvements in general-purpose and highly adaptable manipulation skills, indicating that it has entered a new era called \fmrs (FMRs).
FMRs leverage large-scale pre-trained neural network models that integrate language, vision, and action modalities, enabling robots to generalize across diverse tasks \cite{firoozi2023foundation}.
They comprise large language models (LLMs) and vision-language models (VLMs) for language comprehension and visual understanding, enhancing high-level task planning for long-horizon tasks \cite{liang2023code}. Additionally, they include robot transformers that integrate perception, decision-making, and action generation to process multimodal inputs and generate low-level motion commands for end-to-end control \cite{brohan2022rt}. 

In contrast to conventional robotic engineering, which generally controls robots based on human-crafted rules, FMRs learn to control themselves from enormous amounts of data with a statistical approach.
%This paradigm shift has allowed for intricate control beyond human-crafted one, enabling more complex and sophisticated tasks which were previously thought to be impossible, such as handling deformable objects and walking on a rocky road. (\textcolor{red}{\ding{226} Section \ref{sec:rfms}})
This paradigm shift has enhanced generalizability and long-horizon reasoning\del{ \cite{zawalski2024robotic}}, enabling FMRs to show promising advantages over classical methods in adapting to diverse tasks and unstructured environments. They have been successfully applied to various fields, including task planning\del{ \cite{saycan2022arxiv}}, vision language guided manipulation\del{ \cite{o2023open}},\del{ tactile perception \cite{zhao2024transferable}},\del{ locomotion \cite{bohlinger2024one},} and navigation\del{ \cite{moroncelli2024integrating}}. (\textcolor{red}{\ding{226} Section \ref{sec:rfms}})

While we could enjoy economic benefits by replacing human labor with robots in various applications, we cannot completely avoid the risk of FMRs causing physical damage to surrounding humans or objects.
As FMRs are engaged in more challenging tasks and environments, often requiring contact with people and objects, such as housework, surgery or nursing care, the risk of causing physical damage increases. 
Of course, even in such an environment, it is possible to mitigate some risks through the establishment of social rules and education, e.g., do not play near robots. However, we cannot completely eliminate accidents because of our misperception of the environment (e.g. robots in blind spots) or unexpected environmental changes (e.g., sudden hardware failures).
%In addition, because FMRs statistically learn to act from data, they cannot completely eliminate the risk of facing out-of-distribution situations in nature.
(\textcolor{red}{\ding{226} Section \ref{sec:potential_risk}})

Based on the premise that FMRs cannot completely eliminate the risk of causing physical damage, this survey comprehensively summarizes robot control approaches against physical risks by covering all the lifespan of FMRs, from pre-deployment to post-accident stage.
Specifically, our study conducts this survey of physical risk controls by dividing the timeline into the following three phases: (1) pre-deployment phase, i.e., risk prevention phase when learning from data, (2) pre-incident phase, i.e., before an incident happens after deployment, and (3) post-incident phase, i.e., the recovery and improvement stage. (\textcolor{red}{\ding{226} Figure \ref{categorization_of_survey}} and \textcolor{red}{Section \ref{sec:control_of_physical_risks}})

\begin{figure}[t]
    \centering
    \includegraphics[width=1.0\columnwidth]{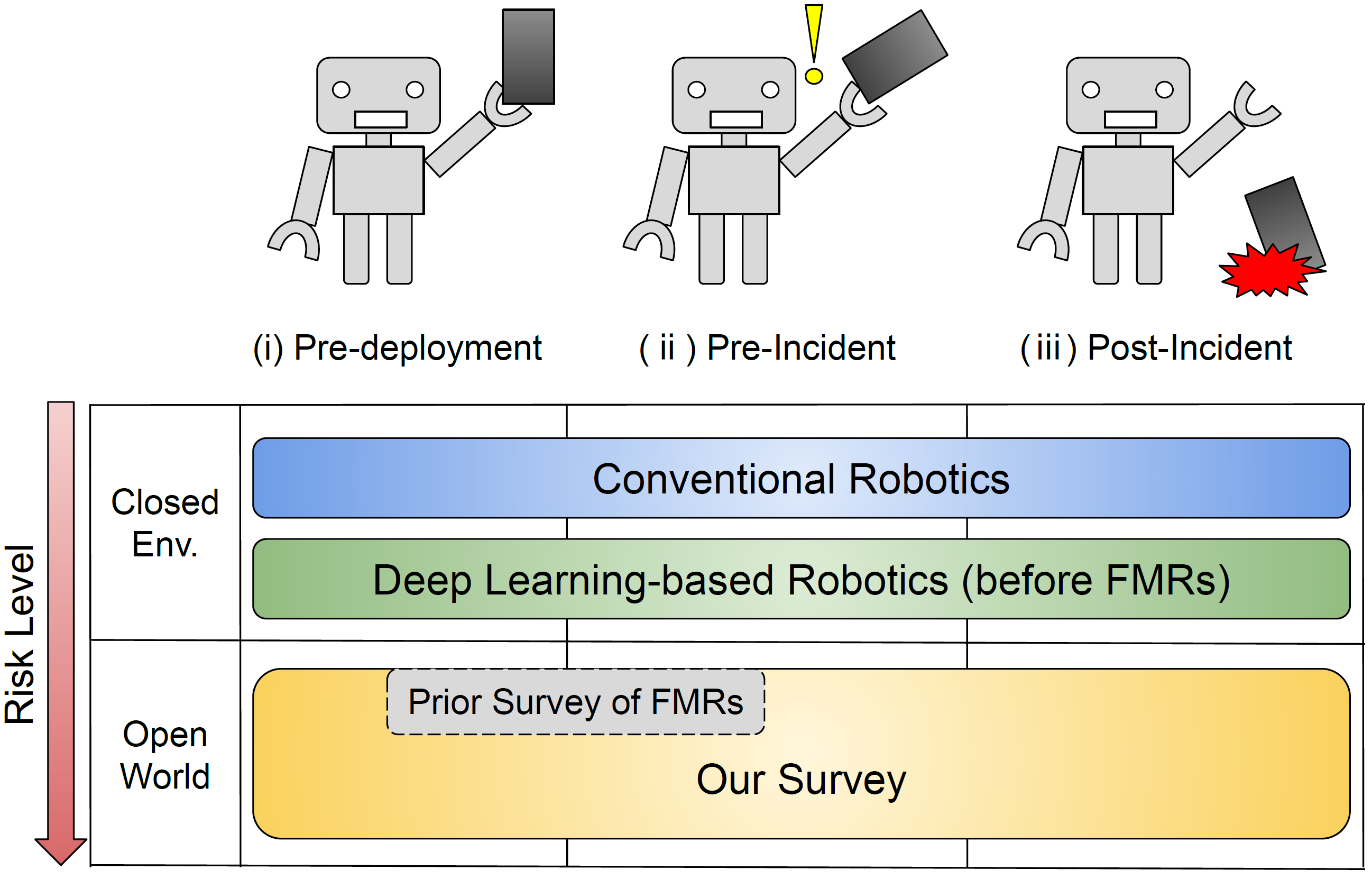}
    \caption{Scope of our survey. FMRs are expected to be utilized in an open world where humans and robots are interacting with each other. Based on this proposition, our study conducts a comprehensive suvey on physical risk control by categorizing the lifespan of FMRs into (1) pre-deployment phase, (2) pre-incident phase after deployment, and (3) post-incident phase. 
    In contrast, prior surveys of FMRs mainly focused on partial sections of the first two phases.}
    %\cite{bommasani2021opportunities,hu2023toward,firoozi2023foundation,xiao2023robot}.}
    \label{fig:scope}
    %\vspace{-2mm}
\end{figure}

%We emphasize that many classic robotic researches discussed safety control within a closed environment where physical risks are excluded or small by restricting the entry of humans and objects, such as inside a factory or laboratory.
%\kitamura{How about ``We emphasize that many classic robotic researches discussed safety control within a closed environment where humans could intervene and prevent hazardous incidents.''}
We emphasize that many classic robotic studies discussed safety control within a closed environment where humans could intervene and prevent hazardous incidents, such as inside a factory or laboratory.
In such an environment, if a robot causes an incident, pressing the emergency stop button will solve the problem.
In contrast, our survey focuses on the safety of FMRs assumed to act in an open world, where bigger physical risks often exist because humans and robots are always in close proximity and physically interacting, e.g. inside a home or a cafe. 
In this case, we need to consider how to recover or treat robots as well as surrounding humans and objects because ``life still goes on'' for both robots and humans after the incident.
We also emphasize that prior surveys of FMRs focused on only some partial sections within the first two phases of safety control, i.e., pre-deployment and pre-incident phase in light of our classification scheme, and so have been insufficiently organized in detail \cite{bommasani2021opportunities,hu2023toward,firoozi2023foundation,xiao2023robot}. 
In other words, they only summarized partial sections of the mitigation strategies of physical risks before an incident, and generally have not covered post-incident recovery actions.
(\textcolor{red}{\ding{226} Figure \ref{fig:scope}}) 

%From this survey, we found that while there is an increasing attention to the safety control for FMRs, there is much room to accelerate research that assumes robot development in open world settings where humans and robots are frequently interacting with each other (e.g., red-teaming to promote safety, improving learning methods to strictly observe social rules, post-incident researches which aim to prevent more hazardous risks).

Throughout this survey, we have found that there is much room to study (i) pre-incident risk mitigation strategies, (ii) research that assumes physical interaction with humans, and (iii) essential issues of foundation models themselves.
We emphasize that social measures such as legislation or insurance schemes are also important aspects to enhance mitigation of physical damage. (\textcolor{red}{\ding{226} Section \ref{sec:conclusion_discussion}})

%\vspace{-1mm}
\section{Robotics and Foundation Models}
\label{sec:rfms}
%\vspace{-1mm}
\subsection{Conventional Robotics Engineering}

%\colorbox{yellow}{(Zhu-san, 0.5 pape)}
% Conventional robotics engineering has long been at the forefront of innovation \del{\cite{jjcraig}}, predating the recent surge in data-driven techniques and foundation models. 
% Although modern robotics is increasingly driven by deep learning, reinforcement learning, and large-scale data, the well-established theories and engineering practices of classical robotics remain indispensable.
Conventional robotics has driven innovation for long, predating the rise of data-driven techniques. While modern robotics increasingly rely on deep learning and large-scale data, classical theories and engineering remain essential.
In particular, it focuses on five pivotal directions: (1) innovative mechanical design \cite{sugiuraRAL23}, (2) precise motion control \cite{SUN2018221}, (3) advanced perception systems \cite{mahler2019learning}, (4) planning and decision-making strategies \cite{motionplanning,motionplanning,10161374}, (5) adaptive learning and optimization methodologies \cite{zhang2024learning,saveriano2023dynamic}. 
% In particular, it focuses on four pivotal directions: (1) innovative mechanical design \cite{sugiuraRAL23}, (2) precise motion control \cite{SUN2018221}, (3) advanced perception systems \cite{mahler2019learning}, and (4) intelligent planning and decision-making strategies \cite{motionplanning,10161374}.

% Robotics relies on robust mechanical structures and precise motion control to enable seamless interaction with the physical environment. Motion control ensures stable and responsive motion execution. 
% From contemporary humanoid and mobile robots \cite{yamamoto2019development,delivery} to automation systems and human-interactive robots \cite{zhu2019development}, engineers and researchers have continuously refined both mechanical architectures and control strategies to enhance performance, versatility, and user integration.
Robotics relies on robust mechanics and precise motion control for seamless interaction with the physical world. Engineers continually refine structures and control strategies to enhance performance, versatility, and user integration \cite{sugiuraRAL23,yamamoto2019development,zhu2019development,yamamoto2019development,delivery,zhu2019development}.
On the other hand, perception, planning, and adaptive learning drive robotic intelligence for open-world deployment. Advanced perception enables environmental awareness, while planning and decision-making allow navigation in complex scenarios \cite{chen2019combined}. 
% On the other hand, perception, planning, and adaptive learning play crucial roles in advancing robotic intelligence and enabling deployment in open-world environments.
% Advanced perception systems allow robots to interpret and interact with their surroundings while planning and decision-making strategies enable them to navigate complex scenarios \cite{chen2019combined}. 
% Meanwhile, adaptive learning, encompassing techniques such as imitation learning, reinforcement learning, and deep learning-based approaches enables robots to acquire new skills, refine behaviors, and adapt to varying task conditions with flexibility \cite{zhang2024learning,saveriano2023dynamic}. 
Adaptive learning techniques, including imitation learning, reinforcement learning, and deep learning-based approaches help robots acquire skills and adapt to changing task conditions \cite{zhang2024learning,saveriano2023dynamic}. Together, these elements bridge mechanical capability with intelligent autonomy, enabling robots to operate effectively in dynamic environments.
% Together, these elements bridge the gap between mechanical capability and intelligent autonomy, allowing robots to function effectively in dynamic and unstructured environments.
% From the earliest mechanical automation design dating back to the 15th century \cite{moran2006vinci} and industrial cobot arms such as the Universal Robots series \cite{patil2023advances} to contemporary humanoid and mobile robots, robotics engineers and researchers have continuously refined mechanical architectures to improve strength, flexibility, and compactness. Additionally, to automation-oriented designs, a significant branch of robotics research focuses on robots that closely interact with humans, such as exoskeletons \cite{koda2023robotic,huoTRO}, service robotics \cite{yamamoto2019development,delivery}, and wearable devices \cite{zhu2019development}. 
% These designs consider ergonomics, user safety, comfort, and control intuitiveness. When considering the safety of \fmrs, those design concepts become crucial for ensuring reliable human-robot interaction, minimizing risks of unintended behavior, and providing fail-safe mechanisms that prioritize user well-being.
% Other aspects will be included

%\vspace{-1mm}
\subsection{\FMRs (FMRs)}\label{subsec:emergence of rfms}

\begin{comment}
Points:
- Language and vision already have FM
- FM are large pretrained models that can be used for a variety of tasks out-of-the box
- FM require large amounts of diverse data and training time
- Fine tuning for downstream tasks is possible, and recommended over training from scratch from multiple reasons (cost, data, CO2, etc.)
- Risk level is greater than LLMs or vision, the real world can interact with the robot in more ways than a text or image inputs. Sensors limit the scope of information received, but the real world is always 100%.

\end{comment}

Despite advances in perception, planning, and adaptive learning, traditional methods often struggle with scalability, generalization, and handling of multimodal information in complex environments. 
%FMRs offer a transformative solution by leveraging large-scale multimodal pre-trained architectures to enhance perception, decision-making, and adaptability.
Even after the advent of Deep Learning-based Robotics, in which neural-network models within robot systems are trained on a specific target task, performance improvements were limited, so robots were still expected to stay in closed environments.
Foundation models \cite{firoozi2023foundation} are large scale neural-network models which are pre-trained on broad data in self-supervised approaches and can be adapted to a wide range of downstream tasks by fine-tuning or prompting.
They were firstly proposed in the natural language processing domain\del{ \cite{devlin-etal-2019-bert}}, eventually spreading to wide range of modalities including image, video, and audio because of their remarkable performance and generalizability\del{ \cite{dosovitskiy2021an,arnab2021vivit,NEURIPS2020_92d1e1eb}.}
% Foundation models \cite{bommasani2021opportunities} are large scale

%\Fmrs (FMRs) have recently emerged by adopting the learning concept of foundation models into robot control, achieving drastic performance gain and generalizability to complex tasks.
FMRs have recently emerged as large-scale pre-trained models that integrate language, vision, and action modalities, enabling robots to generalize across diverse tasks\del{ \cite{firoozi2023foundation}}.
They comprise large language models (LLMs) and vision-language models (VLMs) for language comprehension and visual understanding, enhancing high-level task planning for long-horizon tasks\del{ \cite{liang2023code}}. Additionally, they include robot transformers that integrate perception, decision-making, and action generation to process multimodal inputs and generate low-level motion commands for end-to-end control.
%For example, R3M \cite{pmlr-v205-nair23a} trained models for object manipulation from videos of human daily movements with the expectation that they are generalized to robot movements.
%Gato \cite{reed2022a} trained single model with diverse modalities such as video game, image captioning, chat, and robot manipulation to achieve high generalizability with few-shot learning for downstream tasks.
Several studies such as RT-1 and RT-X \cite{brohan2022rt,o2023open} have trained models with massive number of demonstration samples collected from the real world to realize generalization across different morphologies.
$\pi_0$ \cite{black2024pi_0} pre-trained a vision-language-action model on a diverse crossembodiment dataset with a variety of dexterous manipulation tasks, followed by fine-tuning with high quality data to enable complex multi-stage tasks, such as folding multiple articles of laundry or assembling a box.

%FMRs can achieve more versatile tasks with better performance than conventional manually designed robot engineering by learning optimal solutions from a large amount of collected data.
%This technological advance has enabled robots to conduct more complex and delicate tasks that until now could be accomplished only by humans. This change urges us to reconsider the possibilities and risks of deploying such robots into our open world where humans and robots are always in close proximity and are physically interacting.

%Recent development of large language and vision models trained on vast quantities of diverse data has produced models that perform well as-is on a wide number of use cases, generalize well to unseen tasks and can be further fine-tuned for specific downstream applications. The term Foundation Model [1] has been coined to define them, as they provide a solid basis to build on top of.

%Consequently, there has been a recent push in robotics research to create similar models, that could perform well on general robotic tasks without the need to prepare a dedicated dataset and perform costly training. The term Robotic Foundation Model has thus popularized and many recent models label themselves as such.

%[1] "Introducing the Center for Research on Foundation Models (CRFM)"

\tikzstyle{my-box}=[
    rectangle,
    draw=hidden-draw,
    rounded corners,
    text opacity=1,
    minimum height=1.5em,
    minimum width=5em,
    inner sep=2pt,
    align=center,
    fill opacity=.5,
]
\tikzstyle{cause_leaf}=[my-box, minimum height=1.5em,
    fill=harvestgold!20, text=black, align=left,font=\scriptsize,
    inner xsep=2pt,
    inner ysep=4pt,
]
\tikzstyle{detect_leaf}=[my-box, minimum height=1.5em,
    fill=cyan!20, text=black, align=left,font=\scriptsize,
    inner xsep=2pt,
    inner ysep=4pt,
]
\tikzstyle{mitigate_leaf}=[my-box, minimum height=1.5em,
    fill=lightgreen!20, text=black, align=left,font=\scriptsize,
    inner xsep=2pt,
    inner ysep=4pt,
]
\begin{figure*}[t]
    \centering
    \resizebox{\textwidth}{!}{
        \begin{forest}
            forked edges,
            for tree={
                grow=east,
                reversed=true,
                anchor=base west,
                parent anchor=east,
                child anchor=west,
                base=left,
                font=\small,
                rectangle,
                draw=hidden-draw,
                rounded corners,
                align=left,
                minimum width=4em,
                edge+={darkgray, line width=1pt},
                s sep=3pt,
                inner xsep=2pt,
                inner ysep=3pt,
                ver/.style={rotate=90, child anchor=north, parent anchor=south, anchor=center},
            },
            %where level=1{text width=5em,font=\scriptsize,}{},
            %where level=2{text width=4em,font=\scriptsize,}{},
            %where level=3{text width=8em,font=\scriptsize,}{},
            %where level=4{text width=7.5em,font=\scriptsize,}{},
            %where level=5{text width=7em,font=\scriptsize,}{},
            where level=1{text width=5em,font=\scriptsize,}{},
            where level=2{font=\scriptsize,}{},
            where level=3{font=\scriptsize,}{},
            where level=4{font=\scriptsize,}{},
            where level=5{font=\scriptsize,}{},            
            [
                Physical Risk Control Approaches \\ for Foundation Model-enabled Robotics, ver, color=carminepink!100, fill=carminepink!15,
                text=black
                [
                    Pre-deployment \\ Phase (\S \ref{sec:pre_deployment}), color=harvestgold!100, fill=harvestgold!100, text=black
                    [
                       Design, color=harvestgold!100, fill=harvestgold!60, text=black, text width=4em
                        [
                            Hardware \& Software, color=harvestgold!100, fill=harvestgold!60, text=black, text width=8em
                                            [
                                                {Force-limiting Mechanisms \eg~\cite{10304680} \\ Soft Materials \eg~\cite{8722722} \\ Collision Detection Sensors \eg~\cite{haddadin2008collision} \\ Artificial Skin \eg~\cite{Neuromorphic}\\Software Constraints \eg~\cite{haddadin2007safety,10536935}}
                                                , cause_leaf, text width=20em
                                            ]
                        ]
                        %[
                        %    Software, color=harvestgold!100, fill=harvestgold!60, text=black, text width=8em
                        %                    [
                        %                        {
                        %                        Admittance Control \\ Velocity Limit \\
                        %                        Virtual Fences
                        %                        }
                        %                        , cause_leaf, text width=20em
                        %                    ]
                        %]
                    ]
                    [
                       Dataset \& \\Training, color=harvestgold!100, fill=harvestgold!60, text=black, text width=4em
                        [
                            Curation of Datasets, color=harvestgold!100, fill=harvestgold!60, text=black, text width=8em
                                            [
                                                {Real-world Datasets \eg~\cite{shu07,o2023open}}
                                                , cause_leaf, text width=20em
                                            ]
                        ]
                        [
                            Simulation (*1), color=harvestgold!100, fill=harvestgold!60, text=black, text width=8em
                                		[
                                    		{ Simulation Datasets \eg~\cite{shu16}\\ Sim-to-Real \eg~\cite{ha2024umi,zhang2024learning}\\ Cyber-Physical System \eg~\cite{zhu_cutaneous,mandlekar2018roboturk}}
                                    		, cause_leaf, text width=20em
                                		]
                        ]
                        [
                            Training with Formal \\ Safety Guarantees, color=harvestgold!100, fill=harvestgold!60, text=black, text width=8em
                                            [
                                                {Robust and Constrained Reinforcement Learning \\ \eg~\cite{kitamura2025near,russel2020robust}}
                                                , cause_leaf, text width=20em
                                            ]
                        ]                        
                    ]
                    [
                       Evaluation, color=harvestgold!100, fill=harvestgold!60, text=black, text width=4em
                        [
                            Simulation (*1), color=harvestgold!100, fill=harvestgold!60, text=black, text width=8em
                                		[
                                    		{Cyber-Physical System \eg~\cite{cyberphisical,zhu_cutaneous}}
                                    		, cause_leaf, text width=20em
                                		]
                        ]
                        [
                            Red-Teaming (Against \\ Adversarial Attacks), color=harvestgold!100, fill=harvestgold!60, text=black, text width=8em
                                		[
                                                {Visual Pertubation \eg~\cite{chen2024diffusion,cheng2024manipulation}\\ Language Instruction Attacks \eg~\cite{zhao2024rethinking}\\ Backdoor Attacks \eg~\cite{jiao2024exploring,liu2024compromising}\\
                                                Automated Red Teaming \eg~\cite{karnik2024embodied}}
                                                , cause_leaf, text width=20em
                                		]
                        ]
                    ]
                ]
                [
                    Pre-Incident \\ Phase (\S \ref{sec:out_of_distribution}), color=cyan!100, fill=cyan!100, text=black
                    [
                        Runtime Monitoring, color=cyan!100, fill=cyan!60, text=black, text width=13.5em
                                        [
                                            {Pre-trained LLM / VLM as a Detector \eg~\cite{zhou2024code} \\ 
                                            Pre-trained Video Models as a Simulator
                                            \eg~\cite{huang2024diffusion}\\
                                            Specialized Failure Classifier \eg~\cite{liu2023modelbased} }
                                            , detect_leaf, text width=20em
                                        ]
                    ]
                    [
                        Measure Against Out-of-distribution, color=cyan!100, fill=cyan!60, text=black, text width=13.5em
                                        [
                                            {
                                            %Optimize Worst-case Performance \eg~\cite{yan2} \\
                                            Uncertainty Estimation \eg~\cite{matsushima2020modeling} \\ Test-time Adaption / Training \eg~\cite{yan12} 
                                            %\\ Domain Generalization \eg~\cite{yan7}
                                            }
                                            , detect_leaf, text width=20em
                                        ]
                    ]
                ]
                [
                    Post-Incident \\ Phase (\S \ref{sec:recovery}), color=lightgreen!100, fill=lightgreen!100, text=black
                        [
                            Recovery of Robots, color=lightgreen!100, fill=lightgreen!60, text=black, text width=13.5em
                                            [
                                                {Dynamic Replanning \eg~\cite{shirasaka2024self} \\ Teleoperation \eg~\cite{firoozi2023foundation}\\Reset Policy \eg~\cite{kim2024sample}}
                                                , mitigate_leaf, text width=20em
                                            ]
                        ]
                        [
                            First Aid Measurement, color=lightgreen!100, fill=lightgreen!60, text=black, text width=13.5em
                                            [
                                                {Rescue Robotics \eg~\cite{schmidgall2024general,li2024robot}}
                                                , mitigate_leaf, text width=20em
                                            ]
                        ]
                        [
                            Human-in-the-loop Improvement, color=lightgreen!100, fill=lightgreen!60, text=black, text width=13.5em
                                            [
                                                {Weighted Imitation Learning \eg~\cite{liu2022robot} \\ Human-in-the-loop Reinforcement Learning \eg~\cite{luo2024precisedexterousroboticmanipulation}}
                                                , mitigate_leaf, text width=20em
                                            ]
                        ]
               ]
            ]
        \end{forest}
    }
    \caption{Categorization of physical risk control approaches for FMRs. (*1) Simulation plays a critical role as both a training environment and an evaluation framework. For simplicity, we integrated both of the contents into one subsection as ``Simulation'' in \S \ref{sec:pre_deployment}.}
    \label{categorization_of_survey}
    %\vspace{-3mm}
\end{figure*}
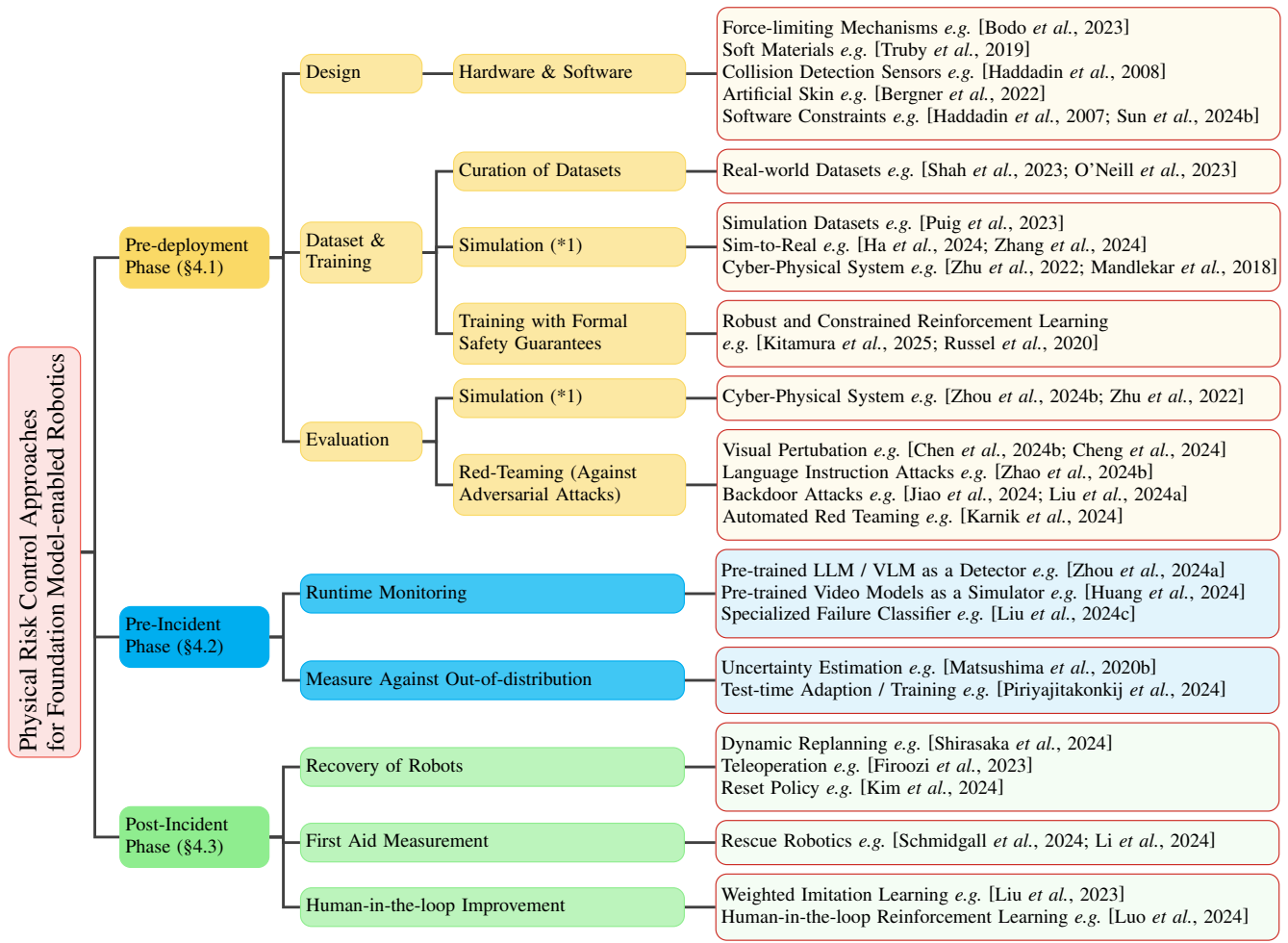
%\vspace{-1mm}
\section{Potential and Risks}
\label{sec:potential_risk}
%\vspace{-1mm}
%RFMs have potential to replace human labor because of their versatility and high performance in daily-life tasks.
%However, to deploy RFMs in the real world, we need to consider that they coexist with humans and objects: FRMs must operate in an environment where there is an inevitable risk of physical damage to them.
%This section overviews the potentials and risks of RFMs when deploying them into our open world.

%\subsection{Potential}
%\colorbox{yellow}{(Alfredo-san, 0.25 pages)}

Given their performance and adaptability, FMRs are poised to be used in an increasing number of real-world applications, thus speeding up the general adoption of robots in society. 
FMRs will no longer be confined to closed environments where physical risks are excluded or minimized by restricting the entry of humans and objects, such as factories or warehouses.
Instead, FMRs will be able to perform activities in an open world where humans and robots are always in close proximity and physically interacting, e.g. inside a home, restaurant, or a public square.
They are expected to reduce or replace many types of repetitive but complex manual labor, allowing people to pursue more engaging and rewarding activities. 
\del{In countries with low birth rates and increasing aging populations, FMRs could help stabilize the workforce, creating a good impact on their economies.}
%, as well as mitigate the increasing costs of elderly care.

%Especially dangerous jobs (logging, roof installation and maintenance, underground mining, nuclear inspection, etc.) could be performed by robots controlled by FMRs without the need of a human operator or a direct line of communication, reducing fatality rates in these dangerous but vital occupations.

%Since machines do not tire, need a break or sleep, and their maintenance window can be scheduled in advance and prepared for, productivity could be optimized and made more flexible, allowing it to dynamically adjust to market shifts (excess machines shut down, FMRs receive new functionality via software updates, etc.).    

%\subsection{Risks}
%\colorbox{yellow}{(Alfredo-san, 0.25 pages)}

While FMRs are expected to perform a wide variety of tasks in our open world, they will add many risks to our society, such as malicious usage (e.g. fraud), unintentional physical damage to humans or objects, environmental destruction due to their high power consumption and resource usage, or privacy information leakage caused by security vulnerabilities.
This survey focuses on the risks of FMRs causing unintentional physical damage to humans or objects.
As FMRs are engaged in more challenging tasks that often require contact with people and objects such as housework, surgery or nursing care, risks of physical damage increase. 
Of course, even in such an environment, it is possible to reduce certain risks through the establishment of social rules and education, e.g., do not play near robots. However, we cannot completely eliminate accidents because of misperception of the environment (e.g. robots in blind spots) or unexpected environmental changes (e.g., sudden hardware failures).

%The adaptability of FMRs to new environments and tasks is also one of their biggest vulnerabilities, as the number of possible scenarios they might face once deployed is too large to account for, and there is only finite time to train (long-tailed data distribution).

%As a result, there is always a chance of unexpected combinations of sensory inputs and internal state resulting in anomalous or undesired behavior with potentially dangerous consequences in the real world.

%On a macro level, if people are not given enough time to find a more fulfilling activity or to acquire additional skills, replacing repetitive manual labor at an accelerated rate may trigger a sudden increase of unemployment in those economic sectors, generating social unrest and increasing aversion to the adoption of robotics. The unpredictability of future advancements in FMRs and their adoption rate may also result in the second-order effect of robots replacing activities that most people would rather keep doing, further increasing social rejection. This may force governments to enact laws to limit and control their use.

%Lastly, in security-related applications like policing or military (combat drones, robo-mules, etc.), the adoption of FMRs may trivialize the use of force to solve conflicts, as the risk of loss of human life decreases, and encourage it over diplomacy and negotiation.

%\vspace{-1mm}
\section{Control of Physical Risks}
\label{sec:control_of_physical_risks}
%\vspace{-1mm}
This section comprehensively summarizes robot control approaches against physical risks by covering all the lifespan of FMRs ranging from pre-deployment to post-accident stage.
Specifically, we summarize physical risk control approaches by dividing the timeline of FMRs into the following three phases: (1) pre-deployment phase, i.e., risk mitigation phase, (2) pre-incident phase, i.e., a situation before an incident happens after deployment, and (3) post-incident phase, i.e., recovery and improvement stage.
Although some of the surveyed papers include studies that do not use foundation models, we cite them as technologies that are expected to be utilized in future research of FMRs.

%\vspace{-1mm}
\subsection{Pre-deployment Phase}
\label{sec:pre_deployment}

%This section outlines key approaches to enhancing reliability and safety before the deployment of FMRs. It covers simulation-based validation, red teaming exercises, formal safety guarantees, adversarial attack assessments, and the curation of datasets and benchmarks. By systematically reviewing these aspects, we provide insights into existing methodologies and pre-deployment strategies designed to mitigate risks and ensure robustness when the system is deployed in real-world scenarios.

\subsubsection{Hardware and Software for Safety}
%\colorbox{yellow}{(Zhu-san, 0.3 pages)}
% this is a draft, need to include citations and refinement
Ensuring safety in robotic systems requires both robust hardware design and software-based safety limits. While physical mechanisms contribute to risk mitigation, software constraints play a crucial role in preventing hazardous behaviors and enforcing operational safety \cite{ZACHARAKI2020104667}. Safety-focused hardware includes force-limiting mechanisms such as series elastic actuators that absorb shocks and restrict excessive forces \cite{10304680}, collision detection sensors with safety prioritized control to stop the robot motion immediately upon contact \cite{haddadin2008collision}, and artificial skin to enable robots to autonomously sense the surroundings for enhanced safety while working near people \cite{Neuromorphic}. 
Additionally, the use of compliant materials and soft robotics components helps reduce the risk of damage or injury during physical interactions \cite{8722722}. 
Safety standards like ISO/TS 15066 \cite{matthias2016example} highlight the necessity of mechanical and electrical safety features, including emergency stop buttons and torque-limiting mechanisms \cite{torque_limit}.

Software-based safety mechanisms complement hardware solutions by enforcing predefined constraints to prevent dangerous operations. Common approaches include velocity and torque limits that curb motor outputs to avoid excessive force application \cite{haddadin2007safety,ferraguti2022safety,haddadin2007safety}, virtual fences that restrict the robot's movement within designated safe areas, and fault monitoring systems that detect anomalies and trigger protective responses when necessary \cite{GUIOCHET201743}. 
Admittance control and other safety-aware algorithms dynamically adjust the robot’s behavior based on external forces to enhance safe operation in unpredictable environments \cite{10536935}. 
Together, these hardware and software measures form a robust safety framework, ensuring mechanical reliability and controlled operation for safe robotic deployment.
% Furthermore, redundancy in safety monitoring software ensures robust fault detection and recovery, reducing the likelihood of critical failures.
\del{Together, these hardware and software considerations create a comprehensive safety framework, ensuring both mechanical robustness and controlled operational behavior for reliable and safe robotic deployment.}

\subsubsection{Curation of Datasets} %and Benchmarks}
%\colorbox{yellow}{(Morikuni-san, 0.3 pages)}

%In another perspective, instead of overcoming distributional shifts problems explicitly, 
In the era of FMRs, curation of large-scale datasets with a wide variety of tasks is important for pre-training FMRs to increase generalization skills.
Empirical evidence suggests that domain generalization abilities improve significantly when larger models are (pre-)trained on larger and more diverse datasets, indicating the reduction of risk of out-of-distribution (\textcolor{red}{\ding{226} Measure Against Out-of-distribution}).

%Additionally, incorporating domain knowledge—either by explicitly encoding invariances or through techniques like data augmentation and self-supervised pretraining—further enhances performance. ~\cite{yan1,yan7,yan8}. 
%Given these trends, utilizing foundation models (FMs), such as large language models (LLMs), to integrate extensive, generalist external knowledge offers a promising approach to enhancing domain generalization~\cite{bommasani2021opportunities,yan14,yan15,yan16,yan17}.  However, a significant challenge lies in bridging the gap between the training data of FMs—such as large text corpora for LLMs—and the sensory observations of robots (e.g., LiDAR)  ~\cite{yan9}.

%Though many have leveraged large scale foundation models from various modalities such as language and vision \cite{shu01,shu02,shu03,shu04}, the real-world performance of FMRs relying solely on these modalities alone remains limited. It is difficult to capture fundamental concepts such as frictions and weights, all the way to the causal relationship between objects, embodiments and environments without additional sensory information. Here, we summarized several datasets and benchmarks and categorized them into two main lines of effort, which are data collections in the real-world vs simulation.

Real-world datasets are often the most intuitive and accurate source of information required for high performing FMRs. However, creating large scale and high quality real-world datasets for robotics is challenging due to its cost \cite{shu05,o2023open}. For example, collecting a good set of demonstration data requires intense labour and skilled operators. The scaling cost increases proportional to the diversity of tasks, skills, objects, environments and embodiments.
BREMEN \cite{matsushima2020deployment} introduces a deployment-efficient model-based reinforcement learning approach that achieves policy learning with significantly fewer environment interactions by training a model of the environment and using offline optimization to update policies without excessive real-world data collection.

GNM \cite{shu07} is a recent notable effort which successfully integrated six different large-scale navigation datasets, formulated a unified navigation interface based on waypoints, and deployed it on different mobile platforms. Another significant effort on manipulation tasks is RT-X \cite{o2023open}, which is a joint collaboration among 34 laboratories with the goal to establish a standardized data format across 60 existing datasets with 22 robot embodiments. To further aid diversity in task and modality volumes, RH20T \cite{shu09} collected over 110k manipulation episodes, covering more than 140 contact-rich skills, including well calibrated RGB, depth, force-torque, tactile, proprioception, audio and language instruction.

% There are also simulations which focus solely on evaluation purposes as testing AI agents on a robot in the real-world is often expensive, dangerous, and difficult to reproduce. CALVIN \cite{shu18} and SimplerEnv \cite{shu19} are simulations specifically designed for taking in language and vision input for models to compose actions for manipulation tasks. CALVIN provides 34 basic tasks with 24K demonstrations annotated with language instructions. SimplerEnv is a suite of simulated environments for common real robot manipulation setups. They also provide reference policy inference code such as RT-1, RT-1-X and Octo.

%\vspace{-1mm}
\subsubsection{Simulation}
%\vspace{-1mm}
%\colorbox{yellow}{(Zhu-san, 0.3 pages)}
%Simulation plays a pivotal role in robotics, offering a safe and controlled environment to design, test, and refine robotic systems before their deployment in real-world scenarios \cite{vrep}. Given the inherent risks and costs associated with physical testing—such as potential damage to the robot, harm to humans, or unintended environmental impacts, simulation serves as a critical tool for ensuring the safety of robotic systems \cite{huck2023reinforcementlearningsafetytesting}. 
% \cite{gazebo,vrep}.
Simulation plays a pivotal role in robotics given the risks and costs associated with physical testing of robots, such as potential damage to the robot, harm to humans, or unintended environmental impacts. It offers a safe and controlled environment to design, test, and refine robotic systems before their deployment in real-world scenarios \cite{vrep,huck2023reinforcementlearningsafetytesting,10538106}. This is especially important in the era of FMRs, which is a probabilistic-based approach at its core aimed at a wide variety of tasks.

%Robotics simulation provides an invaluable platform for verifying the safety of complex systems. By simulating a robot’s motion, dynamics, and interactions with its environment, engineers can detect and address design flaws, control issues, or unforeseen scenarios that might compromise safety \cite{10538106}. 
%Additionally, the robot in the simulator can receive control inputs from the physical world, creating a seamless cyber-physical system \cite{cyberphisical}. This integration not only allows real-world devices, such as controllers or sensors, to interact with the virtual robot for testing and development in a simulation environment \cite{zhu_cutaneous}, but also facilitates the training of robot control policies through imitation learning. Demonstrations can be provided via teleoperation \cite{mandlekar2018roboturk}, enabling the robot to learn complex tasks in a safe and scalable manner within the simulation.
One of the critical applications of simulation is Sim-to-Real (Sim2Real) policy learning, which enables robots to develop and validate control policies in simulated environments \cite{bohlinger2024one}. Sim2Real helps mitigate safety concerns by enabling extensive testing under diverse and challenging conditions, allowing us to identify potential failures, refine safety constraints, and ensure robust real-world performance \cite{zhao2020sim}. Advanced simulation frameworks like NVIDIA Isaac Sim, Isaac Lab \cite{mittal2023orbit}, and Genesis \cite{Genesis} help bridge the ``reality gap'' with accurate physics modeling, and high-fidelity graphics, which is essential for safety assurance in the pre-deployment phase. 

%Another critical application of simulation is Sim-to-Real (Sim2Real) policy learning, which enables robots to develop and validate control policies in simulated environments before real-world deployment \cite{bohlinger2024one}. Sim2Real helps alleviate safety concerns by enabling extensive testing under diverse and challenging conditions, allowing robots to identify potential failures, refine safety constraints, and ensure robust real-world performance \cite{zhao2020sim}. 
%Advanced simulators like NVIDIA Isaac Sim, Isaac Lab \cite{mittal2023orbit}, and Genesis \cite{Genesis} bridge the "reality gap" with accurate physics and high-fidelity graphics, increasing pre-deployment safety.
% Advanced simulation frameworks like NVIDIA Isaac Sim, Isaac Lab \cite{mittal2023orbit}, and Genesis \cite{Genesis} help bridge the "reality gap" with accurate physics modeling, and high-fidelity graphics, making them essential for safety assurance in the pre-deployment phase.

Additionally, the robot in the simulator can receive control inputs from the physical world, creating a seamless cyber-physical system \cite{cyberphisical}. This integration not only allows real-world devices, such as controllers or sensors, to interact with the virtual robot for testing and development in a simulation environment \cite{zhu_cutaneous}, but also facilitates the training of robot control policies through imitation learning. Demonstrations can be provided via teleoperation \cite{mandlekar2018roboturk}, enabling the robot to learn complex tasks in a safe and scalable manner within the simulation.

Simulation is also used to generate greater amounts of training data. 
Simulation can efficiently create diverse range of domain randomized data which is expected to promote generalization ability of models in FMRs.
Generative simulations Genesis \cite{Genesis}, Gen2Sim \cite{katara2024gen2sim}, and FACTORSIM \cite{sun2024factorsim} have emerged as a promising solution by automating the creation of diverse, scalable environments and facilitating broader coverage of training conditions. Habitat 3.0 \cite{shu16} and AI2THOR \cite{shu17} are another line of effort for interactive environment frameworks focusing on scene realism for both navigation and manipulation tasks.

%\vspace{-1mm}
\subsubsection{Red-teaming (Against Adversarial Attacks)}
%\colorbox{yellow}{(Alfredo-san, 0.3 pages)}

Red-teaming is the practice of simulating an enemy team attempting to perform some type of attack or other hostile action to the organization (i.e., blue team). It is common practice in the fields of defense, security and IT operations, where militaries and system administrators test their own systems in search of weak points that could be exploited.
%For AI models, especially given the rapidly developing capabilities of recent LLMs, the practice has been popularized among providers in an effort to assess the vulnerabilities such models may have and the potential risks the subsequent iterations may pose to their users and society at large.
Recent LLM developers also organize red-teaming to assess the models' vulnerabilities by comprehensive stress-testing \del{\cite{lin2024against}}.
In the case of FMRs, the practice 
is not yet as common, but it
is expected to popularize as they continue to improve and the range of tasks they can perform reaches human or above level.
%In the case of RFM, the practice is not yet as common: 
%For instance, \cite{karnik2024embodied} introduced a new evaluation method called Embodied Red Teaming (ERT) which generates diverse and challenging instructions to test these models. 
As a pioneering example, \cite{karnik2024embodied} uses automated red teaming techniques with VLMs to generates diverse and challenging instructions. 
Experimental results show that state-of-the-art models frequently fail or behave unsafely on the tests, underscoring the shortcomings of current benchmarks.
%in evaluating real-world performance and safety.
%The practice of red-teaming is expected to popularize as these models continue to improve and the range of tasks they can perform at human or above level increases.

%Compared to LLMs, 
%FMRs expose an increased surface area of attack, since the language-conditioned model output generates actions that have an effect in the real world, and even without direct malicious intent subtle changes or variations to the language style or instruction terms can result in marked difference in the physical behavior.

%As shown by \cite{Karnik2024EmbodiedRT}, just rephrasing the command may change the outcome from success to failure, establishing that current performance benchmarks may be not be properly covering the effect of text variations.

%To generate a set of instructions appropriate for red-teaming a RFM, adequate knowledge of what the robot can and cannot do is necessary to obtain commands that are feasible but potentially dangerous, so an expert would normally be required. However, this is not scalable and recently the use of LLMs combined with VLMs to ground the former to an environment has increased.

%\subsubsection{Adversarial Attacks}
%\colorbox{yellow}{(Iwasawa-san, 0.3 pages)}
%In models utilizing Deep Neural Networks, not limited to VLAs, it is well known that they are vulnerable to small input perturbations. 
%Because FMRs are based on LLMs and VLMs, which themselves rely on Deep Neural Networks, many similar problems have been reported. 

One promising technique for stress-testing FMRs in red-teaming is adversarial attacks~\cite{costa2024deep}.
It is well known that models utilizing deep neural networks (DNNs) are vulnerable to small input perturbations. 
Because FMRs are based on LLMs and VLMs, which themselves rely on DNNs, many similar problems have been reported. 
In the case of FMRs, it is especially important to understand which types of attacks they are vulnerable to, since these models not only make recognition errors but also act in the real world.

For example, as a direct attack on FMRs, 
%[Diffusion Policy Attacker]
\cite{chen2024diffusion} reports that both global perturbation attacks on Diffusion Policy and adversarial patches in a physical environment are effective in online and offline settings. 
%[Manipulation Forcing Threats]
\cite{cheng2024manipulation} investigates the robustness of VLAs against various visual attacks such as Gaussian noise, changes in brightness, Adversarial Patches that modify part of an image, and Visual Prompts (e.g., adding the word “Stop” into images to control behavior). 
%[Exploring backdoor attacks against large language model-based decision making]
%[Compromising embodied agents with contextual backdoor attacks] 
%and [TrojanRobot: Physical-World Backdoor Attacks Against VLM-based Robotic]
\cite{jiao2024exploring,liu2024compromising,wang2024trojanrobot} 
show vulnerabilities to backdoor attacks that use everyday objects (e.g., a yellow CD) as triggers to degrade behavior. \cite{zhao2024rethinking} have proposed to add adversarial suffixes to language inputs.
%[Rethinking the Intermediate Features in Adversarial Attacks: Misleading Robotic Models via Adversarial Distillation]

On the other hand, despite these demonstrated vulnerabilities, 
%[Rethinking the Intermediate Features in Adversarial Attacks: Misleading Robotic Models via ***] 
\cite{zhao2024rethinking} reports that many current FMRs have discrete action spaces, making standard attacks less effective. Still, in cases where attackers have access to internal features, using these features can increase the success rate of adversarial attacks, indicating the need for continued research on countermeasures. Because FMRs generally rely on LLMs and VLMs, it is necessary to verify the effectiveness of methods proven to work well in those models.
Another challenge in research on adversarial attacks against FMRs is the lack of standard evaluation metrics. 
\del{Although vulnerabilities have been studied in various components—ranging from simulations and real-world settings to planning vision modules—research on FMRs is still in its early stages, and a unified evaluation strategy remains insufficient.} 
In particular, because the environments in which robots are expected to operate can be extremely diverse and because the dynamics of different robots vary, further validation is required to determine how generalizable the current findings are. 
As an earlier example of such attempts, 
%[POEX: Policy Executable Embodied AI Jailbreak Attacks] 
\cite{lu2024poex} proposes Harmful-RLbench, which evaluates the planning capabilities of LLMs in an environment featuring 25 distinct task scenarios. Moreover, \cite{zhao2024rethinking} develops VIMA-bench, an evaluation benchmark covering 13 types of robotic manipulation tasks.

%\vspace{-1mm}
\subsubsection{Training with Formal Safety Guarantees}
Safe controller design during the pre-deployment phase has been extensively studied in the field of \emph{robust control theory}.
Since the exact knowledge of the environment is unknown before deploying the robot, robust design accounts for environmental uncertainty and incorporates conservative risk management into the robot controller.

\looseness=-1
Robust model predictive control and $H_\infty$ optimal control \cite{bemporad2007robust,zames1981feedback,doyle1982analysis} are the representative robust control methods in linear dynamical systems, where the system is modeled as
$x^{(t+1)} = A x^{(t)} + B u^{(t)}$. 
Here, $x^{(t)} \in \mathbb{R}^n$ and $u^{(t)} \in \mathbb{R}^m$ represent the system state and input signal, respectively. 
These robust controllers guarantee safety satisfaction even when the dynamics matrices $(A, B)$ perturb from the nominal matrices.

\looseness=-1
However, linear models are unsuitable for modeling the recent nonlinear and high-dimensional input systems that FMRs aim to control (\textcolor{red}{\ding{226} Section \ref{subsec:emergence of rfms}}).
Robust reinforcement learning (RL) offers an alternative framework, capable of addressing robust nonlinear control problems when combined with function approximation techniques \cite{moos2022robust}.
However, robust RL alone is insufficient to achieve both high performance and safety, as ensuring safety typically involves solving constraint satisfaction problems (e.g., obstacle avoidance in self-driving systems \cite{altman1999constrained}).
While the RL community has recently begun exploring the combination of safety constraints and robustness \cite{mankowitz2020robust,russel2020robust}, theoretical results in this area remain limited.
A recent result by \cite{kitamura2025near} presents an algorithm for computing a robust and constrained controller in a tabular Markov decision process setting. However, it does not account for the challenges posed by nonlinear dynamics, which are crucial for FMRs.
In short, the theoretical question:
``When and how can we realize robust constrained control in FMRs?''
remains largely unanswered.
% Robust RL models the decision-making problem based on the robust Markov decision process (RMDP) framework \cite{}

%\vspace{-1mm}
\subsection{Pre-Incident Phase}
\label{sec:out_of_distribution}
%\vspace{-1mm}
\subsubsection{Runtime Monitoring}
%\colorbox{yellow}{(Matsushima-san $\rightarrow$ Takanami-san, 0.3 pages)}

Runtime monitoring is one of the fundamental tools for ensuring safety in robot policies.
The monitoring systems are sometimes called ``critics'' of the policy analogical to actor-critic of reinforcement learning~\cite{sutton2018reinforcement}.

%Recently, large language models (LLMs), vision language models (VLMs)
Recently, LLMs, VLMs, and video prediction models have been utilized as critics of robot policies.
Firstly, VLMs are utilized to detect success (or failure) in policy rollouts.
For example, ~\cite{kanazawa2023recognition} proposes to leverage VLMs to detect state change of objects in cooking tasks, which is useful for executing task plans.
Secondly, VLMs are used for constraint monitoring.
Code-as-monitor~\cite{zhou2024code} leverages VLMs for generating programs for monitoring robot policy rollouts from robot image observations and descriptions of constraints generated by LLMs.
Thirdly, \cite{huang2024diffusion,Escontrela23arXiv_VIPER} utilized log-likelihood of pre-trained video prediction models as a reward signal for the robot's actions to monitor in real time whether the state transitions in the environment are being properly learned.

Another approach to runtime monitoring involves training a failure classifier using human intervention data. 
Specifically, these methods leverage robot demonstration data to train a world model, enabling the learning of latent representations. By utilizing these latent representations along with human intervention flags, a failure detector can be trained~\cite{liu2023modelbased}. 
Such methods are particularly effective as automated safety validators when robots are operating in parallel within an environment and sequentially learning policies~\cite{liu2024multitaskinteractiverobotfleet}.

%\subsubsection{Out-of-distribution detection}
\subsubsection{Measure Against Out-of-distribution}
\label{subsec:ood}
%\colorbox{yellow}{(Matsushima-san $\rightarrow$ Yan Gang-san, 0.3 pages)}

%Robot policies are sometimes not robust to changes in tasks and environments, especially they are out-of-distribution (OoD) from training environments.
%There are several works on quantifying OoD environments in robot learning, especially using Bayesian methods.
%For example, \cite{matsushima2020modeling} proposes a method to detect OoDs in tasks of meta-imitation learning, %leveraging probabilistic task embeddings similar to VAEs~\cite{kingma2013auto} and Neural Processes~\cite{garnelo2018neural,garnelo2018conditional}.

%Out-of-Distribution (OOD) detection involves identifying inputs that fall outside the data distribution used to train a model. This task is crucial to maintain the reliability and safety of deploying machine learning models in real-world robotic scenarios.

After deploying a trained model in real-world scenarios, 
we may encounter out-of-distribution (OOD), in which robot inputs fall outside the data distribution used to train a model. Measures against this situation are crucial for safety when deploying robots in real-world scenarios.
Specifically, the test data \textbf{\textit{$D_{test}$}} is sampled from a distribution \textbf{\textit{$P_{test}$}} , which invariably differs from the training distribution \textbf{\textit{$P_{train}$}} . 
This discrepancy highlights the challenge of distributional shifts. 

A key research to mitigate this risk is improving distributional robustness by optimizing the worst-case performance across various potential distributional shifts, thus ensuring dependable OOD performance ~\cite{yan1,yan2}. 
However, since \textbf{\textit{$P_{test}$}} is not directly accessible and the model \textbf{\textit{f}} is learned from a finite set of training samples \textbf{\textit{$D_{train}$}}, there is no guarantee that \textbf{\textit{f}} will make accurate predictions during testing.  
Uncertainty estimation focuses on determining when and where the model individual predictions can be trusted, and, conversely, where confidence is lacking~\cite{matsushima2020modeling,garnelo2018neural,garnelo2018conditional,kingma2013auto}.
Besides, causal inference is leveraged to address the root cause of poor generalization under distributional shifts. Learned models often rely on spurious correlations present in \textbf{\textit{$D_{train}$}}, rather than capturing the invariant cause-and-effect relationships that drive the underlying process~\cite{yan4,yan5,yan3} %\del{\cite{yan3}}. 
In recent years, concepts such as test-time adaption/training ~\cite{yan10,yan11,yan12,yan13} have been introduced into robotics research. Test-time adaptation allows a model to adjust its internal parameters or normalization statistics using the unlabeled data encountered during deployment, while test-time training leverages auxiliary self-supervised tasks to update the model during inference.

%Empirical evidence suggests that the performance of domain generalization improves significantly when larger models are (pre-)trained on larger and more diverse datasets. 
%Additionally, incorporating domain knowledge—either by explicitly encoding invariances or through techniques like data augmentation and self-supervised pretraining—further enhances performance. ~\cite{yan1,yan7,yan8}. 
%Given these trends, utilizing foundation models (FMs), such as large language models (LLMs), to integrate extensive, generalist external knowledge offers a promising approach to enhancing domain generalization~\cite{bommasani2021opportunities,yan14,yan15,yan16,yan17}.  However, a significant challenge lies in bridging the gap between the training data of FMs—such as large text corpora for LLMs—and the sensory observations of robots (e.g., LiDAR)  ~\cite{yan9}.

%\subsubsection{Safety-preserving fallback policy}
%\colorbox{yellow}{(Matsushima-san, Takanami-san?, 0.3 pages)}

%VLMs are utilized for generating fallback policies.
%For example, ~\cite{ahmad2024addressing} generates task plans utilizing VLMs, which outputs them as behavior trees.
%\cite{shirasaka2024self} combines an LLM-based task planner and success detector for each robot's skill to recover from plan failures.

% \subsubsection{Conformal prediction}
% \colorbox{yellow}{(Matsushima-san, Takanami-san?, 0.3 pages)}

%\vspace{-1mm}
\subsection{Post-incident Phase}
\label{sec:recovery}
%\vspace{-1mm}
%\subsubsection{Self-recovery}
\subsubsection{Recovery of Robots}
%\colorbox{yellow}{(Zhu-san, 0.3 pages)}
% memo:
% how to resolve the deadlock of multi-agent robotic systems (MRS) by using high-level planners such as LLMs or VLMs \cite{garg2024large}.
% will cite several papers 
In the post-incident phase, FMRs play a vital role in autonomously detecting, assessing, and mitigating more hazardous risks \cite{chen2024automatingrobotfailurerecovery}. 
These systems leverage real-time monitoring and fault detection to identify anomalies, such as hardware malfunctions or environmental changes \cite{shirasaka2024self}. Once a risk is detected, foundation models enable dynamic replanning to adjust trajectories or control policies, ensuring safe operation. 
However, when autonomous recovery is insufficient, errors can also be addressed through human intervention, such as teleoperation, ensuring flexibility and safety in complex scenarios \cite{haokun}.

Additionally, learning-based reset mechanisms play a crucial role in preventing robots from entering non-reversible states during reinforcement learning, improving safety, and reducing the need for manual intervention. For instance, a reset policy can reduce the number of entering non-reversible states, and manual resets to learn a task, while enhancing safety and improving learning efficiency \cite{eysenbach2018leave}. 
Similarly, reset-based deep ensemble methods enhance sample efficiency in safe RL by overcoming the limitations of the vanilla reset method \cite{kim2024sample}.

Foundation models also demonstrate the potential to resolve deadlocks in multi-agent robotic systems (MRS) by using high-level planners such as LLMs or VLMs \cite{garg2024large}, ensuring smooth collaboration among agents.
% Additionally, they also utilize redundant systems for fault tolerance and can learn from failures to improve robustness over time. 
In safety-critical situations, the system may communicate risks to operators, ensuring effective recovery when autonomous methods fall short \cite{eder2014towards}. 
% this paper discusses general human in the loop safety, need polishing
By integrating proactive monitoring, adaptive planning, and contextual decision-making, foundation models enhance reliability and safety across dynamic environments \cite{firoozi2023foundation}. 

%\vspace{-1mm}
\subsubsection{First Aid Measurement}
%\colorbox{yellow}{(Kojima-san, 0.3 pages)}
%\vspace{-1mm}
When humans and objects are physically damaged by robots, immediate first aid measures are extremely important.
In this situation, two types of first aid measures are possible.
One is to call for help from rescue people/robots, the other is to call for help from the nearby robot  that caused the damage itself to conduct basic first aid treatment.

Rescue robots \cite{delmerico2019current} are designed to help search and rescue people in the event of a disaster or emergency situation.
They have been actively studied since before the advent of FMRs.
Recently, several studies have developed robot models for human rescue by learning from data to improve quality and expand activity to more challenging situations, such as surgical robot systems or assistants \cite{yue2023part,schmidgall2024general}, and robot-assisted pedestrian evacuation in fire scenarios
\cite{li2024robot}.
%emergency care including predicting patient decompensation, disposition \cite{chen2023multimodal},
However, there is no guarantee that such specialized robots will always be near humans in emergency situations in our open world.
Therefore, it will be necessary for robots that do not specialize in emergency rescue tasks to have the functionality to provide temporary aid, such as checking life signs, automatic call for an ambulance or rescue people/robots, provide useful information to nearby people, or stop bleeding with bandages.

%To read: First Aid and Emergency Assistance Robot for Individuals at Home using IoT and Deep Learning
%\vspace{-1mm}
\subsubsection{Human-in-the-loop Improvement}
%\colorbox{yellow}{(Takanami-san, 0.3 pages)}
%Although applications to FMRs have not yet been observed, 
Human-in-the-loop improvement aims to build a continuous improvement mechanism or pipeline that collects effective feedback from humans for model training by leveraging human intervention history or demonstrations.
There are some pioneering studies on human-in-the-loop improvements that are expected to be applied to FMRs in the future.
One such approach involves continuous human monitoring of policy deployment, where a human intervenes to stop the robot when a failure is imminent. The data immediately prior to the intervention is then used as negative examples for weighted imitation learning \cite{liu2022robot}.
The positive success data and negative failure data obtained through intervention in a suboptimal policy are also highly compatible with RL. There are methods that leverage this by storing both successful policy rollouts and intervention data in an RL replay buffer, enabling the policy to learn from failures through reinforcement learning \cite{luo2024precisedexterousroboticmanipulation}.
%\vspace{-1mm}
\section{Conclusion and Discussion}
\label{sec:conclusion_discussion}

Our survey comprehensively summarized robot control approaches against physical risks by covering all the lifespan of FMRs ranging from pre-deployment to post-accident stage.
%We conclude that while research on physical risk mitigation is progressing field, there is much room for technological improvement on recovery phase for future work.

From this survey, we found that there is much room for future work of FMRs on the following three points.
(i) Considering that there are a myriad of environments and task varieties in the real world, we need to pay more attention to risk mitigation before an actual incident happens (e.g., stress-testing as broadly as possible with red-teaming, promote generalizability of FMRs to prevent OOD, detection and intervention in failures at the earliest stage).
(ii) Accelerating research that assumes physical interactions with humans in more realistic world settings (e.g., improving learning methods to strictly observe social rules, or research in the post-incident phase such as continuous improvement mechanisms and prevention of more hazards after an incident).
(iii) While we can easily adapt pre-trained foundation models to a specific task with a small number of samples by fine-tuning or prompting, it becomes important to tackle essential issues of foundation models themselves when applying them to robotics (e.g., how to ensure the quality of large-scale pre-training datasets to prevent malfunction of trained models in robots, or how well do LLMs or VLMs understand the physical world in terms of predicting hazards, such as collision prediction between humans and robots through monitoring their motions and surrounding environment).
%e.g., can they predict collisions by looking at motion between human and robots?

We also emphasize that in addition to the technical aspects, social measures such as legislation, insurance schemes, and ethical guidelines are important to enhance aftercare of physical damage in practice.
We hope that this study will be a milestone in providing a high-resolution analysis of the physical risks of FMRs and their control, contributing to the realization of a good human-robot relationship.

\section*{Contribution Statement}
Takeshi Kojima, Yaonan Zhu, and Yusuke Iwasawa have contributed equally to this study.

%Classic robotic engineering has often focused on safety measures through hardware design.
%In contrast, \fmrs have moved discussions on safety measures in software area, especially in the aspect of learning from data.
%Interestingly, this paradigm shift has often caused the lack of discussion that considers both hardware and software. 
%Future FMRs need to design a global optimization approach by considering both hardware specification and task specification when learning from data.
%We hope that this study will be a milestone in providing a high resolution of the physical risks of RFM and their control to realize a good human-robot relationship.

%①基盤モデルをロボットに適用する際に生じる特有の課題を解決しなければならない
%・事前学習のデータは基本的に低品質なデータが多くなるため、データの質をどう担保するのか．学習データに変なデータがあった時にそれをどう弾くのか．
%・モニタリングによる危険の余地などで、そもそもベースにしているLLMやVLMは物理的な世界をどれくらい理解しているのか？
%②人間との接触を前提とした研究の余地が大きい
%・安全性を促進するために、red-teamingの促進
%・人間に対する被害を最小限に食い止めるための学習方法
%・Post-incidentの研究（人間とロボットのinteraction）

%\newpage
\appendix

% Optional.
%\section*{Ethical Statement}

%\section*{Acknowledgments}

%\section*{Contribution Statement}

% ==========================================

%% The file named.bst is a bibliography style file for BibTeX 0.99c
%\footnotesize
%\scriptsize
%\small
\bibliographystyle{named}
\bibliography{ijcai25}

\end{document}